# Integrated Deep Learning Framework for Reliable Renewable Energy Supply Forecasting

Haibo Wang[1]
A. R. Sanchez, Jr. School of Business, Texas A&M International University, Laredo, Texas, USA

Wendy Wang
School of Computing, University of South Alabama, Mobile, Alabama, USA

Lutfu S. Sua
Decision Sciences and Supply Chain, Southern University and A&M College, Baton Rouge, Louisiana, USA

Jun Huang
Management and Marketing Department, Angelo State University, San Angelo, Texas, USA

Bahram Alidaee
School of Business Administration, The University of Mississippi, University, Mississippi, USA

**Abstract**
Due to the inherent variability of renewable energy sources, predicting renewable energy production requires robust approaches and prediction methods that can detect and derive complex patterns. Since deep learning (DL) models can capture complex, nonlinear relationships, they are preferred over traditional machine learning (ML) methods in the renewable energy sector. An examination of the extant literature suggests the need for comprehensive studies on key factors influencing the accuracy of DL techniques in predicting renewable energy production. We aim to provide more insights in this area. In this study, we evaluate seven machine learning methods - Long Short-Term Memory (LSTM), Stacked LSTM, Convolutional Neural Networks (CNN), CNN-LSTM, Deep Neural Networks (DNNs), Time-Distributed Multilayer Perceptron (TD-MLP), and Autoencoder (AE) - using a dataset combining weather and photovoltaic power output data from 12 locations in Spain. Four regularization techniques - early stopping, neuron dropout, and Lasso and Ridge (L1/L2) regularization - are also applied to address the overfitting issue commonly seen in DL models. The evaluation results demonstrate that with a more extensive training set, the combination of early stopping, neuron dropout, and L1 regularization provides the best performance in reducing overfitting problems in CNN and TD-MLP models, whereas with a smaller training set, the combination of early stopping, neuron dropout, and L2 regularization is most effective in reducing the overfitting issue in CNN-LSTM and AE models.

---

[1] Address correspondence to correspondence to Haibo Wang, Ph.D., Division of International Business and Technology Studies, A. R. Sanchez, Jr. School of Business, Texas A&M International University, 5201 University Boulevard, Laredo, Texas, USA. Email: hwang@tamiu.edu

# 1. Introduction

The rapid growth of renewable energy-generated electricity reflects society's increasing environmental awareness (Ang et al. 2022; Aslam et al. 2021; B. 2023). Integrating renewable electricity into the traditional electrical grid requires reliable predictions of cost and production. Such predictions enable developers and investors to quantify economic benefits, secure project financing, and make informed decisions; they can also enhance grid resilience through optimized maintenance, improve power production forecasting, and ensure system stability. Due to the influence of geo-locations and constantly changing weather conditions, renewable electricity production is precarious and challenging to predict. For instance, solar energy is influenced by sunlight exposure, while wind energy relies on wind speeds and location (Wang et al. 2019).

Machine learning (ML) technologies such as Neural Networks, Time Series Analysis, Ensemble Methods, Tree-based Methods, and Deep Learning (DL) have been applied to analyze historical data to produce more accurate energy output estimates. Among the different ML approaches applied, DL has demonstrated remarkable capabilities in identifying complex patterns in renewable energy-generated electricity data. However, complex models such as DL have been shown to have the issue of overfitting, impacting its generalization to unseen data. The effectiveness and widespread adoption of DL in the energy domain have faced other significant challenges, such as data scarcity, model interpretability, computational demands, generalization to unseen data, vulnerability to adversarial attacks and distribution shifts, etc. These issues demand innovative solutions to ensure the reliability and accessibility of DL in critical applications. Overcoming these obstacles requires hybrid approaches, access to high-quality datasets, and advanced model architecture. Furthermore, regularization must be applied to address DL overfitting by adding a penalty term to the loss function.

These challenges prompt us to investigate the following questions in this study:

(1) RQ1: How do different DL architectures compare in terms of their susceptibility to overfitting?
(2) RQ2: Are there architecture-specific regularization techniques that outperform general methods for certain types of neural networks?

This study proposes an analytics framework integrating various DL algorithms with and without regularization approaches. The proposed framework aims to identify critical factors affecting the reliability and availability of renewable energy output forecasts. It combines DL with sampling techniques to mitigate methodology-driven bias, a common limitation of existing algorithms. The study also employs four regularization approaches to address overfitting in DL models and analyzes the trade-off between overfitting and accuracy. The proposed framework effectively captures the nonlinear relationships between energy production and various factors, including weather and seasonality. To our knowledge, this is the first study to evaluate multiple DL models in the domain of renewable energy output forecasts and examine various regularization approaches in tackling the overfitting issue of DL models.

The study employs a diverse array of deep learning models, including Recurrent Neural Network (RNN)-Long Short-Term Memory (LSTM), Stacked LSTM, Convolutional Neural Networks (CNN), CNN-LSTM, Deep Neural Networks (DNN), Time-Distributed Multilayer Perceptron (TD-MLP), and Autoencoder (AE), each chosen for its specific strengths in handling sequential data, spatial hierarchies, or complex pattern recognition (Kong et al. 2019; Wang et al. 2023; Chan et al. 2023; Oluleye, Chan, and Antwi-Afari 2023).

The remainder of this paper proceeds with a review of DL's applications in the renewable energy sector, followed by an overview of DL methods and experimental analyses comparing various DL techniques, and concludes with findings and future research directions.

## 2. Literature review

ML methods have proven effective in energy system planning, reliability, and security. Deep learning, a subset of ML, has gained prominence in adoption. DL methods are widely adopted in the renewable energy industry for infrastructure design, demand forecasting, anomaly detection, failure prediction, production forecasting, and distribution network optimization (Ang et al. 2022; Aslam et al. 2021; Ying et al. 2023; Bansal 2022; Khan et al. 2022; Sharifzadeh, Sikinioti-Lock, and Shah 2019). These applications utilize weather conditions, satellite data, production and consumption data, and expected market prices. By applying multilayered neural networks (NN) to model complex patterns in datasets, DL can capture nonlinear patterns and adapt them to evolving datasets. Its ability to learn behavioral patterns and detect anomalies makes it particularly suitable for complex problems such as the prediction of renewable energy production.

Renewable energy systems, particularly solar and wind, require accurate forecasting models to manage variability. DL's NN-based models have been developed to predict energy outputs over short timescales, enabling more reliable grid integration (Rangel-Martinez, Nigam, and Ricardez-Sandoval 2021; Syed et al. 2021). DL models have been successfully applied to predict solar power generation, demonstrating their large-scale data processing capabilities and accuracy under variable weather conditions (Chang, Bai, and Hsu 2021; Phan et al. 2023). Convolutional and recurrent NN have effectively predicted solar energy output using weather and historical data, outperforming traditional statistical methods in computational efficiency and accuracy (Kong et al. 2019; Arora et al. 2023).

Because of the different techniques applied, comparative analyses of forecasting techniques are crucial for evaluating algorithm efficacy. Studies have explored quantile estimation of renewable energy production using deep neural networks (DNN) to predict regional outputs (Alcántara, Galván, and Aler 2023). Such investigations highlight the importance of model choice and customization based on specific energy systems and datasets.

Despite DL's remarkable success, challenges such as computational requirements, data quality, and model interpretability persist. Researchers have proposed novel frameworks incorporating data preprocessing and postprocessing techniques to address these issues. Hybrid models such as CNN-LSTM frameworks have shown promise in capturing spatiotemporal dependencies in energy data for photovoltaic power forecasting (Phan et al. 2023; Agga et al. 2022).

### *2.1 DL applications in solar, wind, and tidal renewable energy*

#### *2.1.1 Solar energy*
Solar energy has become a leading source added to the grid due to its abundant availability and decreasing installation costs (Kabir et al. 2018). The global capacity of solar power applications supports development in both energy supply and the labor market (Ortega et al. 2020). As reliance on solar energy increases, accurate output prediction becomes essential. Zhang et al. (2018) utilized deep learning techniques for solar energy production forecasting, showing that DL models outperform traditional ML methods in capturing nonlinear patterns. Similarly, Phan et al. (2023) proposed a novel forecasting framework that integrates data preprocessing and postprocessing to enhance prediction performance. Asghar et al. (2023) introduced a demand-side management approach based on machine learning, highlighting how DL techniques optimize energy consumption alongside production forecasting.

### 2.1.2 Wind and tidal energy

Despite its uncertain nature, wind power benefits from DL applications focus on increasing reliability through wind behavior prediction (Wang et al. 2021). Several studies have compared the efficacy of DL-based approaches in improving forecast accuracy. Wind energy forecasting accuracy relies on successfully predicting wind speed and power generation patterns (Hu et al. 2024). Chen et al. (2020) utilized heterogeneous feature learning to improve forecast accuracy in their deep end-to-end framework for multi-step wind power prediction. To increase the reliability of short-term wind forecasting, Du (2019) suggested an ensemble ML approach, combining local weather station data with numerical weather prediction (NWP) outputs. Furthermore, to improve forecasting models, Khodayar and Wang (2019) presented a spatiotemporal graph deep neural network that efficiently captures spatial correlations in wind data. Comparing different DL models is essential to identify the best methods for forecasting renewable energy. A comparison study on wind turbine power curve monitoring using data mining techniques offered insights into various model performances (Schlechtingen, Santos, and Achiche 2013).

In addition to solar and wind energy, DL is also utilized to predict other types of renewable energies, such as tidal power, offering a reliable alternative for growing energy needs. DNN approaches have been used to forecast design values of tidal power plants based on stream regimes (Shadmani et al. 2023).

### 2.1.3 Hybrid approach

Hybrid renewable energy systems present more significant challenges due to their unpredictable nature, leading to increased interest in DL applications (Zahraee, Assadi, and Saidur 2016). Researchers have also introduced leading performance indicators to compare machine learning workflows in energy research, evaluating their application in harvesting, storing, and converting energy (Yao et al. 2023). Additionally, advancements in Artificial Neural Network (ANN) have been applied to hydrogen production research (Thirunavukkarasu, Sawle, and Lala 2023).

Recent advances in DL have focused on hybrid architectures, combining multiple DL methods to improve prediction performance. A recurrent neural network (RNN) for short-term residential load forecasting has been applied, demonstrating its capability to model sequential dependencies (Kong et al. 2019). A clustering-based DL approach for short-term load forecasting in smart grids with integration of consumption pattern recognition has shown enhanced model robustness (Syed et al. 2021). LSTMs and CNNs have also been applied to predict energy consumption, further illustrating the advantages of hybrid DL models (Abraham et al. 2022).

### 2.2 Overfitting and regularization strategies

Overfitting remains a critical challenge in DL applications for renewable energy forecasting. Various studies have addressed this issue through regularization techniques. Wang et al. discussed ML-based methods for sustainable energy systems, highlighting the importance of model generalization (Wang et al. 2023). Percy, Aldeen, and Berry (2018) studied residential demand forecasting with solar-battery systems, proposing data-driven methods to mitigate overfitting. Harrou et al. (2021) demonstrated a soft sensor-based approach to forecasting energy consumption in wastewater treatment plants, incorporating regularization methods to enhance model stability.

Data overfitting is a major problem in DL. To avoid overly complicated models and improve the generalization of the models to unseen datasets, regularization strategies and various techniques are adopted to apply a penalty to the loss function. Regularization strategies such as L1/L2 regularization, neuron dropout, and early stopping are utilized to address overfitting. According to the findings, CNN-LSTM and AE models with smaller training sets benefit the most from early stopping, dropout, and L2 regularization, whereas CNN and TD-MLP models with larger training sets benefit the most from these strategies.

## *2.3 Summary of DL in renewable energy forecasting*

The following models collectively represent a comprehensive approach to addressing various aspects of renewable energy forecasting, from temporal dependencies to spatial correlations and feature extraction (See Table A1 in the Appendix).

While LSTM and its stacked variant excel at modeling sequential dependencies, CNN is more suited for spatial feature extraction, and hybrid models like CNN-LSTM bridge the gap between spatial and temporal modeling. DNNs and TD-MLPs offer versatility and efficiency but lack temporal awareness, while Autoencoders provide unique capabilities for data compression and anomaly detection, albeit with some sensitivity to noise. Each model offers specific strengths and weaknesses, making them suitable for different data types and tasks. This study combines meta-learning and DL for multivariate time series prediction of renewable energy production. Spatial aggregation and decomposition methods have also been suggested to maintain computational feasibility.

## 3. Data source and DL models

The dataset comprises power output data from solar panels installed in 12 cities in Spain over 14 months. It includes 17 features and 21,045 samples, with independent variables such as panel power output, wind speed, date, season, sampling time, location, latitude, longitude, altitude, ambient temperature, humidity, visibility, pressure, and cloud ceiling (Williams and Wagner 2019). This dataset forecasts photovoltaic panel power output (Table A2 in the Appendix).

Skewness and kurtosis values indicate varying asymmetry and tail behavior across variables. The Autoregressive (AR) model is a commonly used statistical ML model that predicts future values in time series analysis, assuming that the data are stationary. However, variables in renewable energy datasets, such as pressure and location, are likely non-stationary, potentially challenging the strict stationarity assumption in time series analysis using autoregressive models. Therefore, DL is preferred over the AR model.

Sample size plays a crucial role in the performance and efficiency of DL models, involving a complex balance between prediction accuracy and computational requirements. Larger training samples improve model generalization and robustness, resulting in better predictive accuracy and reduced overfitting, particularly for complex tasks or high-dimensional data. However, increased sample sizes come with trade-offs, including higher computational resource demands and longer training times. On the other hand, smaller training samples can lead to challenges such as overfitting and biased models due to insufficient data diversity and incomplete representation of the underlying data distribution. Finding the right balance between sample size, model performance, and computational efficiency is crucial for developing effective DL models. To assess the impact of sample size on DL models, 10% to 50% of the sample is selected as the test set, and the number of rows for training, validation, and test sets is provided in Table 1.

**Table 1.** Sample size of training, validation, and test sets.

| Sample Set | 10% | 20% | 30% | 40% | 50% |
|---|---|---|---|---|---|
| Training | 17046 | 13469 | 10312 | 7576 | 5261 |
| Validation | 1894 | 3367 | 4419 | 5051 | 5261 |
| Test | 2105 | 4209 | 6314 | 8418 | 10523 |

Advances in computational power and data availability have significantly improved predictive modeling efficiency. This study employs seven DL models for power generation prediction, chosen for their ability to manage non-linear relationships and high-dimensional data, which are common in time series analysis. Regularization techniques in deep learning are essential to prevent overfitting, improve generalization, and enhance model performance on unseen data. Table 2 lists the four regularization techniques used in this study.

**Table 2.** Notation of models and regularization techniques.

| Model Name | Regularization Techniques |
|---|---|
| Baseline (B1) | None |
| Regularized 1(R1) | Early stopping |
| Regularized 2(R2) | Early stopping and dropout |
| Regularized 3(R3) | Early stopping, dropout, and L1 regularization |
| Regularized 3(R4) | Early stopping, dropout, and L2 regularization |

*3.1 Early stopping*

Early stopping is a simple and effective method where training is halted once the model's performance on a validation set deteriorates, preventing overfitting by selecting the model from the epoch with the best validation performance (Prechelt 1998). Although it is computationally efficient and easy to implement, early stopping requires careful tuning and patience to avoid halting the training process prematurely.

*3.2 Neuron dropout*

Another widely used method is neuron dropout, where a random subset of neurons is "dropped out" during each training iteration, reducing co-adaptation between neurons and encouraging robust feature learning (Srivastava et al. 2014). This stochastic approach is efficient in complex models such as CNN and RNN. However, it increases training time and requires fine-tuning of the dropout rate.

*3.3 L1 regularization*

L1 regularization penalizes the absolute value of model weights by adding a term $\lambda \sum |w|$ to the loss function, encouraging sparsity by forcing some weights to zero (Tibshirani 1996). This method benefits feature selection in high-dimensional datasets but may inadvertently discard valuable information if not carefully calibrated. In contrast, L2 regularization, or ridge regression, penalizes the squared magnitude of weights, expressed as $\lambda \sum w^2$, which prevents large weight values and stabilizes the learning process (Hoerl and Kennard 1970). While L2 regularization does not enforce sparsity, it effectively improves model generalization and is especially useful when all input features contribute meaningfully to the output.

Comparatively, early stopping is the most straightforward and computationally efficient method, though it cannot control model complexity directly. Neuron dropout is highly effective for deep networks but requires additional computational resources and careful hyperparameter tuning. L1 regularization is advantageous when interpretability through feature selection is needed, while L2 regularization is suitable for applications when reducing model complexity without eliminating features is desired. Each technique addresses different aspects of model overfitting and is chosen based on the dataset's specific characteristics and model architecture.

Model performance in this study is measured by a set of metrics, including Root Mean Square Error (RMSE), Mean Squared Error (MSE), Huber Loss, Mean Absolute Error (MAE), Mean Squared Logarithmic Error (MSLE), and R-squared score (Aldajani 2008; Chai and Draxler 2014; Willmott and Matsuura 2005; Draper and Smith 1998; Huber 1992). The formulations of these metrics are presented in the Appendix. These metrics measure the differences between actual and predicted values, guiding model training and evaluation. The TensorFlow library is used to implement all DL models and error metrics in this study.

## 4. Results and discussion

Applying DL techniques to forecast power output from photovoltaic panels at 12 locations in Spain yields similar accuracy and test ratios across methods. Overfitting was observed across the DL baseline models. Table 3 reports the overfitting of seven DL baseline models using error metrics at different sample sizes.

Our analysis shows that the selection of the test set sample size significantly influences overfitting in DL models. With a 20% test set size, five out of seven DL models exhibited reduced overfitting, suggesting that an 80-20 train-test split may offer a good balance between sufficient training data and adequate model evaluation. However, when the test set size was increased to 50%, all baseline models demonstrated clear signs of overfitting, as evidenced by diverging error metrics between training and test sets. This observation underscores the critical role of data sampling in model performance and generalization.

**Table 3.** Overfitting of DL baseline models at different ratios based on error metrics.

| Model | RMSE | MSE | HUBER LOSS | MAE | MSLE |
|---|---|---|---|---|---|
| RNN-LSTM | 10%, 30%, 40%, 50% | 10%, 30%, 40%, 50% | 10%, 30%, 40%, 50% | 10%, 30%, 40%, 50% | 10%, 30%, 40%, 50% |
| Stacked-LSTM | 10%, 30%, 50% | 10%, 30%, 50% | 10%, 30%, 50% | 10%, 30%, 50% | 10%, 30%, 40%, 50% |
| CNN | 10%, 30%, 40%, 50% | 10%, 30%, 40%, 50% | 10%, 30%, 40%, 50% | 10%, 30%, 50% | 10%, 30%, 40%, 50% |
| CNN-LSTM | 30%, 50% | 30%, 50% | 30%, 50% | 30%, 50% | 30%, 50% |
| DNN | 10%, 20%, 30%, 40%, 50% | 10%, 20%, 30%, 40%, 50% | 10%, 20%, 30%, 40%, 50% | 10%, 20%, 30%, 40%, 50% | 10%, 20%, 30%, 40%, 50% |
| TD-MLP | 10%, 20%, 30%, 40%, 50% | 10%, 20%, 30%, 40%, 50% | 10%, 20%, 30%, 40%, 50% | 10%, 20%, 30%, 40%, 50% | 10%, 20%, 30%, 40%, 50% |
| AE | 50% | 50% | 50% | 50% | 30%, 50% |

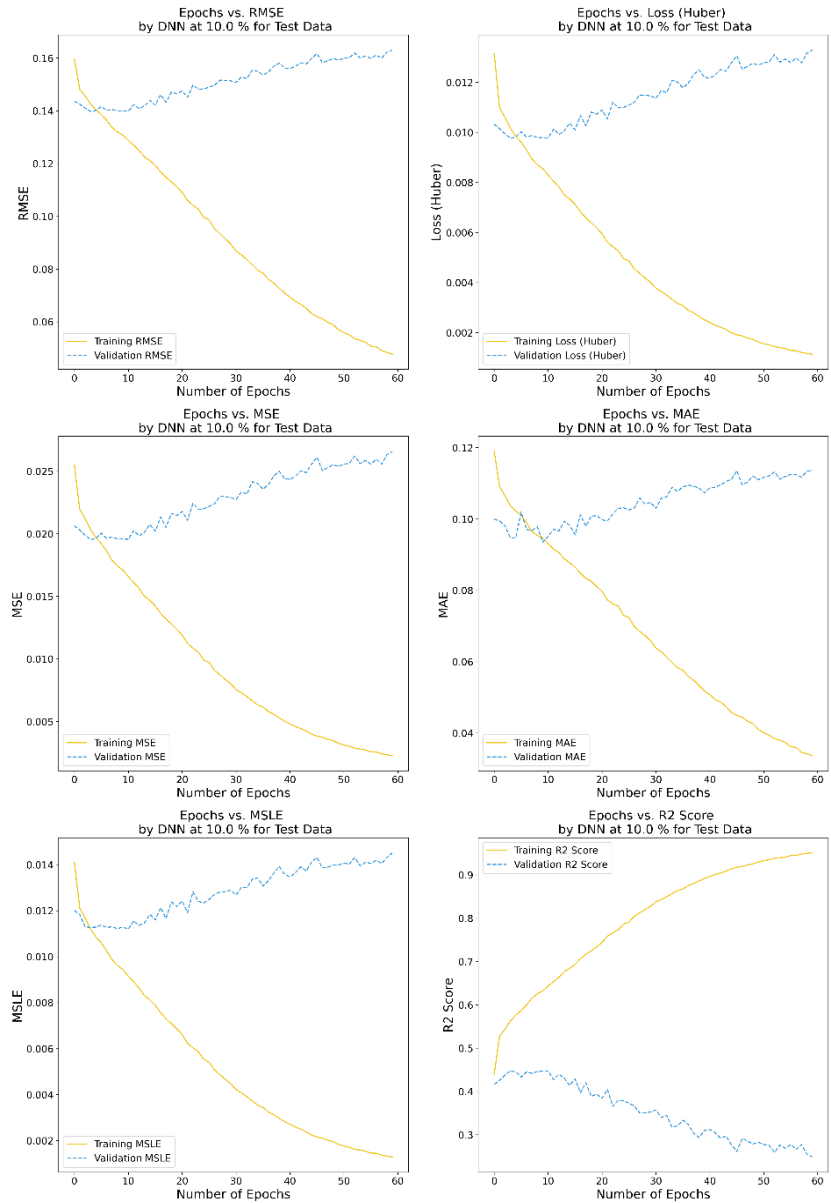

**Figure 1.** Illustration of overfitting in a DNN model with a 10% sample for test data.

The findings in this study highlight the importance of carefully selecting train-test split ratios, as larger test sets can provide more robust evaluation metrics but may compromise model generalization, especially when limited data is available. Furthermore, the varying sensitivities of different DL architectures towards split ratios call for future research to explore the relationship between model complexity, dataset characteristics, and optimal split ratios. The results also underscore the importance of a balanced approach to dataset partitioning, considering both comprehensive model evaluation and effective learning and generalization. For instance, we examine the DNN model at a 10% sample size for overfitting in regard to the best performance in training error metrics. Figure 1 illustrates the learning process for this model, revealing a clear overfitting trend. As the number of epochs increases, the model's accuracy on the training set improves significantly, as evidenced by a consistent decrease in error metrics and an increase in R-squared scores (yellow solid line). On the other hand, the model's accuracy on the validation set deteriorates over time, as indicated by an increase in error metrics and a decrease in R-squared scores (blue dashed line). The gap between training and validation set accuracy widens progressively throughout the learning process, a clear indicator of overfitting as the model becomes increasingly tailored to the training data at the expense of generalization. This overfitting phenomenon becomes more pronounced with increasing epochs, suggesting that early stopping may be an effective regularization technique.

While the DNN model demonstrated the best performance among the seven baseline models, overfitting all models indicates a systemic issue requiring attention. Overfitting becomes more pronounced with a smaller training set and a larger test set as the sample size ratio for the test set increases. The graphical results and detailed reports of different DL models with various ratios and regularization techniques can be accessed from this repository: https://figshare.com/s/381a00d86b9dc42460b5.

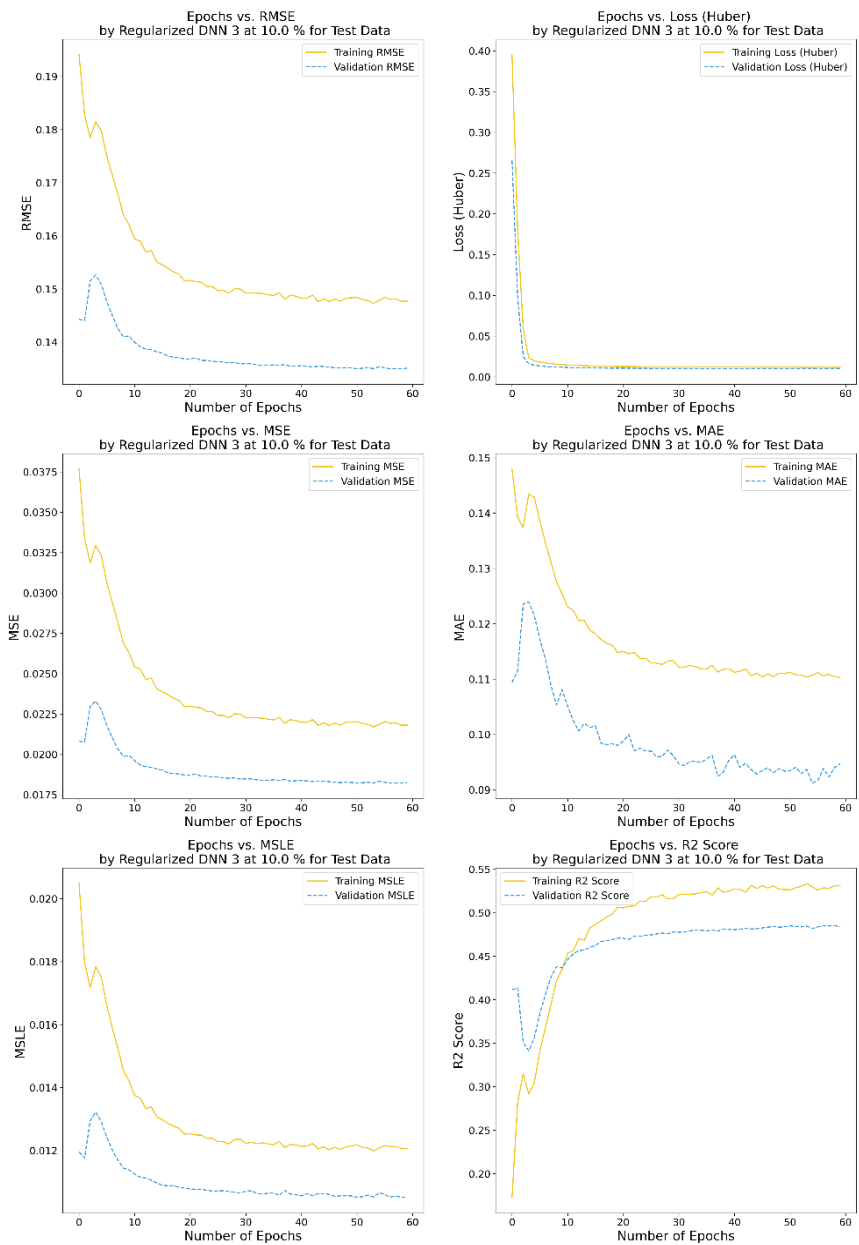

**Figure 2.** Illustration of regularization technique for DNN model with a 10% sample for test data.

In addition to early stopping, we report three additional regularization techniques to address overfitting in DL models. For instance, the combination of early stopping, dropout, and L1 regularization provides the best solution for overfitting in the DNN model, as illustrated in Figure 2.

Table 4 reports the outcomes of regularization techniques in the DNN model as an example. Combining early stopping, dropout, and L1 regularization in R3 offers the best approach to reducing overfitting in DNN models. Early stopping prevents the model from learning noise in the training set by halting training when validation performance degrades, effectively decreasing training time and acting as an implicit regularization technique. Randomly dropping neurons (10% dropout) during training creates multiple sub-networks and reduces co-adaptation. This forces the network to learn more robust features and emulates ensemble learning within a single model. L1 regularization promotes sparsity by forcing some weights to zero, aiding in feature selection and reducing model complexity. These techniques address overfitting from multiple angles: early stopping prevents prolonged exposure to training set noise, dropout introduces beneficial randomness, and L1 regularization simplifies the model structure. Using these techniques together, DNNs can better balance model complexity and generalization ability, resulting in more reliable and robust performance across diverse sample sizes.

**Table 4.** The best regularization technique to reduce overfitting at different ratios based on error metrics.

| Ratio | RMSE | MSE | HUBER LOSS | MAE | MSLE |
|---|---|---|---|---|---|
| 10% | R3 | R3 | R3 | R4 | R3 |
| 20% | R3 | R3 | R3 | R3 | R3 |
| 30% | R3 | R3 | R3 | R3 | R3, R4 |
| 40% | R3 | R3 | R3 | R3 | R3 |
| 50% | R3 | R3 | R3 | R4 | R3 |

Table 4 reports the best performance among the DL models with regularization techniques under different ratios. R3 is the most effective technique for reducing overfitting at lower ratios, and R4 is the best for mitigating overfitting at higher ratios, especially in hybrid models such as CNN-LSTM and AE. R3 is more effective for DL baseline models, such as CNN and MLP, to reduce overfitting, where L1 regularization simplifies the model structure of CNN and MLP by identifying the most critical filters/kernels in CNNs and selecting the most relevant time steps or features in Time-Distributed MLPs. Additionally, for CNNs dealing with high-dimensional image data or Time-Distributed MLPs processing long sequences, L1 regularization can be particularly effective in reducing the impact of irrelevant or redundant information. R4 is more effective on hybrid models such as CNN-LSTM and AE in reducing overfitting, whereas L2 regularization provides a consistent shrinkage of weights. L2 regularization offers significant advantages in addressing overfitting for complex hybrid models like CNN-LSTM and Autoencoders, especially when dealing with limited training data. Adding a penalty term proportional to the square of weights, L2 regularization encourages smaller, more balanced weights across the network, promoting smoother decision boundaries and improved generalization.

This approach is particularly beneficial for the CNN-LSTM architecture, as it helps stabilize LSTM training by mitigating vanishing/exploding gradients and enhancing CNN's ability to learn generalizable spatial features. L2 regularization provides a more distributed representation in the latent space for the AE model. Unlike L1 regularization, L2 maintains all features but reduces their impact proportionally, preserving important spatial and temporal relationships crucial in hybrid models. It also improves robustness to noise and adapts well to features of different scales, making it ideal for processing diverse inputs in the CNN-LSTM model. Furthermore, L2 regularization's compatibility with standard optimization algorithms enhances its practicality in complex architectures. When carefully tuned and combined with other techniques such as dropout or early stopping, L2 regularization provides a powerful tool for balancing model complexity and performance in these hybrid neural network structures.

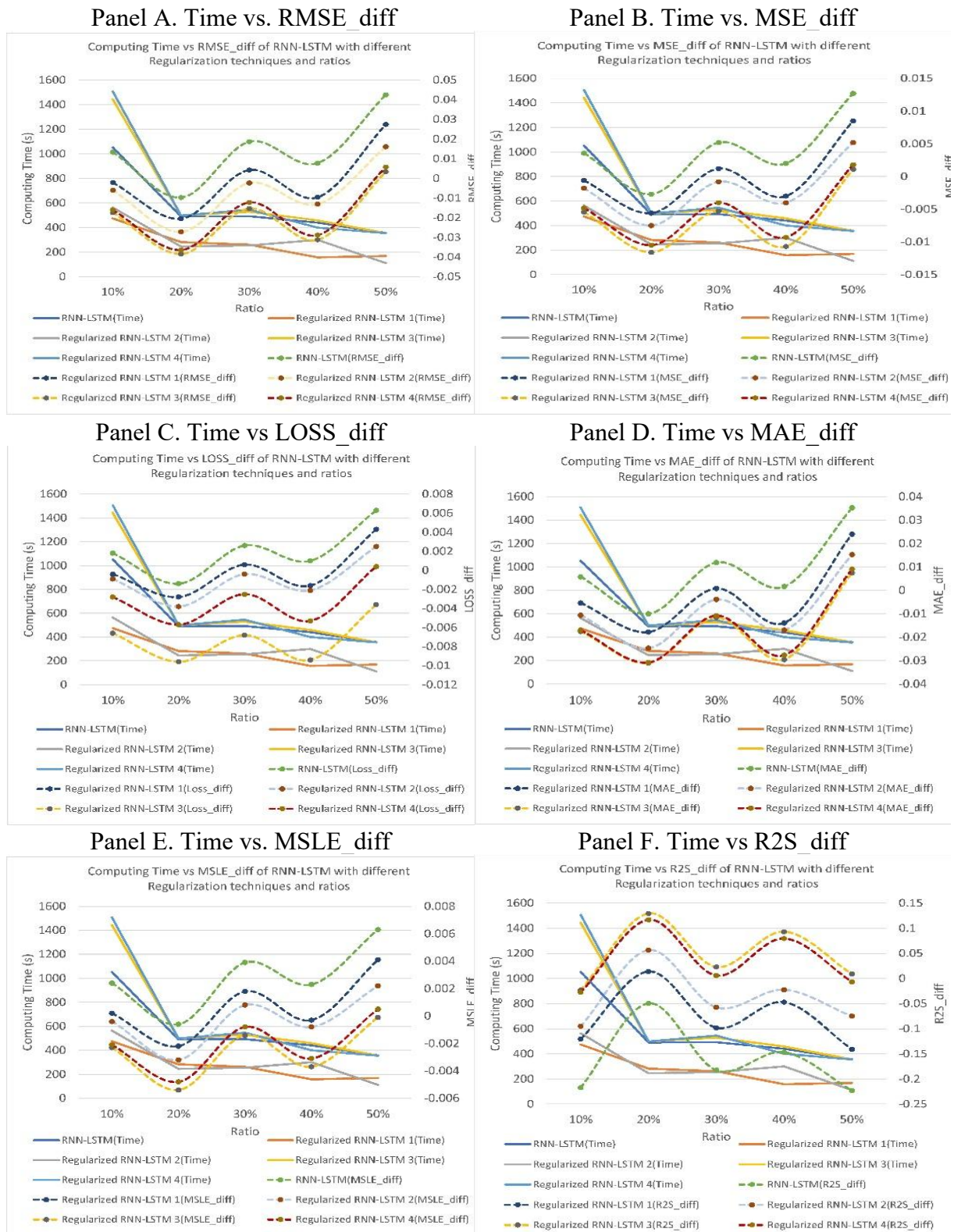

**Figure 3.** Computing time vs. metrics for the RNN-LSTM model with different regularization techniques and sample ratios.

We examine the performance metrics with different sample ratios across regularization techniques for sensitivity analysis and report the findings of the RNN-LSTM model (see Figure 3). Panel A in Figure 3 shows that the RNN-LSTM model achieves the best balance between computing time and the testing/training RMSE difference at a 20% ratio when using Regularization Technique 3, which combines early stopping, dropout, and L1 regularization. A negative RMSE_diff value indicates that the RMSE of the testing dataset is lower than that of the training dataset. We observed similar trends for MSE_diff, LOSS_diff, MAE_diff, and MSLE_diff. Additionally, the highest R2S_diff value at the 20% ratio suggests that the testing dataset achieves a higher $R^2$ score than the training dataset when applying Regularization Technique 3. We find the same results in other models such as the Stacked LSTM, CNN, CNN-LSTM, DNN, TD-MLP, and AE, and our results are reported in https://figshare.com/s/381a00d86b9dc42460b5.

## 5. Conclusion

This study's primary contribution is the comparative application of various DL methods to renewable energy. The regularization techniques on seven DL models were evaluated on a public dataset using five different training/test split ratios, demonstrating their relative performance. This research addresses the need for robust DL methods applicable to multiple renewable energy-related scenarios. Prediction models performed exceptionally well when regularization techniques were used. The test set sample size selection has been shown to significantly address overfitting in DL models. Findings from this study suggest that the combination of early stopping, dropout, and L1 regularization can provide the best performance to reduce overfitting in the CNN and TD-MLP models with a larger training set. In contrast, the combination of early stopping, dropout, and L2 regularization is most effective in reducing overfitting in the CNN-LSTM and AE models with a smaller training set. The study highlights the importance of selecting regularization techniques and DL models tailored to dataset characteristics and prediction tasks.

**Acknowledgment**: This work received support from the Summer Research Award at the A. R. Sanchez, Jr. School of Business and the University Research and Development Awards program at Texas A&M International University.

# Appendix: DL performance metrics

Performance metrics are essential for evaluating model accuracy and generalization across different tasks in machine learning. Each metric provides unique insights into model performance. This appendix includes six major performance metrics used in this study.

Root Mean Squared Error (RMSE) measures the square root of the average squared differences between predicted and actual values. Since this measure penalizes errors more heavily, although it provides interpretable results, it is particularly sensitive to outliers (Chai and Draxler 2014).

$$RMSE = \sqrt{\frac{1}{n}\sum_{i=1}^{n}(y_i - \hat{y}_i)^2} \tag{1}$$

Mean Squared Error (MSE) is widely used as a performance metric in regression models. It is similar to RMSE but without the square root, helping optimize model performance through gradient-based algorithms; however, it shares the same sensitivity to significant errors (Willmott and Matsuura 2005).

$$MSE = \frac{1}{n}\sum_{i=1}^{n}(y_i - \hat{y}_i)^2 \tag{2}$$

Mean Absolute Error (MAE) is another commonly used performance metric that calculates the average absolute difference between actual and predicted values. It is more robust to outliers than MSE. However, it does not penalize significant errors as severely, making it useful when uniform error importance is desired (Willmott and Matsuura 2005).

$$MAE = \frac{1}{n}\sum_{i=1}^{n}|y_i - \hat{y}_i| \tag{3}$$

Huber Loss blends MAE and MSE by employing a quadratic function for small errors and a linear function for large ones, offering robustness against outliers while maintaining the differentiability needed for optimization (Huber 1992).

$$\mathcal{L}_\delta(a) = \begin{cases} \frac{1}{2}(y - \hat{y})^2 & for\ |y - \hat{y}| \leq \delta \\ \delta\left(|y - \hat{y}| - \frac{1}{2}\delta\right) & for\ |y - \hat{y}| > \delta \end{cases} \tag{4}$$

Mean Squared Logarithmic Error (MSLE) evaluates the squared difference between the logarithmic predictions and actual values. It is effective when relative errors are more meaningful, especially in scenarios involving exponential growth or positive-valued predictions (Aldajani 2008).

$$MSLE = \frac{1}{n}\sum_{i=1}^{n}(\log(1 + y_i) - \log(1 + \hat{y}_i))^2 \tag{5}$$

Meanwhile, the R-squared ($R^2$) score, or coefficient of determination, measures the proportion of variance explained by the model, with values ranging from 1, indicating perfect prediction, to negative values, suggesting that the model performs worse than using the mean prediction (Draper and Smith 1998).

$$R^2 = 1 - \frac{\sum_{i=1}^{n}(y_i - \hat{y}_i)^2}{\sum_{i=1}^{n}(y_i - \bar{y})^2} \tag{6}$$

Each of these metrics has specific use cases: RMSE and MSE are ideal for tasks where more significant errors must be penalized more severely, though RMSE is more straightforward to interpret due to its unit consistency. Huber Loss is useful when balancing sensitivity to small errors and robustness against outliers. MAE is preferable when all errors are equally important without emphasizing outliers. MSLE is beneficial when modeling data with wide value ranges or when under-predictions should be penalized more than over-predictions. Finally, the R² score is a useful comparative measure for assessing the model fit across different datasets and scales. The choice of metric depends on the problem characteristics and the desired balance between sensitivity to large deviations, interpretability, and model robustness.

**Table A1.** DL in renewable energy forecasting.

| Names | Approaches | Strengths | Weakness |
|---|---|---|---|
| *Long Short-Term Memory (LSTM)* (Kong et al. 2019; Agga et al. 2022; Abraham et al. 2022) | Capturing long-term dependencies and sequence information in datasets by addressing the vanishing gradient problem through memory cells and gating mechanisms. | Effective for time-series prediction and natural language processing (NLP). | Computationally intensive and slower to train. |
| Stacked LSTM (Kong et al. 2019; Ying et al. 2023; Abraham et al. 2022) | Multi-layered LSTM. | Capturing higher-level temporal features. | Increases model complexity and the risk of overfitting. |
| Convolutional Neural Network (CNN) (Agga et al. 2022; Abraham et al. 2022) | Suitable for spatial data and employing convolutional layers to find local patterns such as edges and textures. | Excel in feature extraction and are robust against noise. | Struggle to capture long-term dependencies. |
| CNN-LSTM (Agga et al. 2022; Abraham et al. 2022) | Local pattern extraction and modeling temporal dependencies. | Combination of strength from both CNN and LSTM. Practical for spatiotemporal data like video analysis. | A hybrid approach, requiring more computational resources and careful tuning. |
| Deep Neural Network (DNN) (Chan et al. 2023; Alcántara, Galván, and Aler 2023) | Fully connected feedforward model with multiple hidden layers. | Offering versatility across various tasks. | Lacking the ability to capture temporal relationships efficiently. |
| Time-distributed MLP (TD-MLP) | Independent application of a multi-layer perceptron to each time step of a sequence while maintaining the sequence's structure. | Allows parallel processing of time-series data. | Cannot directly model temporal dependencies. |
| Autoencoder (AE) | An unsupervised learning model that compresses and reconstructs data through an encoder-decoder framework. | Useful for dimensionality reduction, noise removal, and anomaly detection. | Noise-sensitive and require careful tuning to prevent information loss. |

**Table A2.** Descriptive statistics of the dataset on main variables.

| Variable | Power output (Watt) | Humidity (%) | Ambient temp (C) | Wind speed (km/h) | Visibility (km) | Pressure (millibar) | Cloud ceiling (km) |
|---|---|---|---|---|---|---|---|
| Mean     | 12.9785 | 37.1219 | 29.2851 | 10.3183 | 9.7000  | 925.9447 | 515.9668 |
| Median   | 13.7987 | 33.1237 | 30.2891 | 9       | 10      | 961.1    | 722      |
| Std Dev  | 0.0491  | 0.1642  | 0.0852  | 0.0440  | 0.0093  | 0.5874   | 2.0811   |
| Skewness | -0.0353 | 0.6652  | -0.3264 | 0.6270  | -5.1447 | -0.3588  | -0.8224  |
| Kurtosis | -1.0822 | -0.2626 | 0.16133 | 0.5282  | 27.2766 | -1.5580  | -1.2527  |